\documentclass[lettersize,journal]{IEEEtran}
\usepackage{amsmath,amsfonts}
\usepackage{bm}
\usepackage{algorithmic}
\usepackage{array}
\usepackage[caption=false,font=normalsize,labelfont=sf,textfont=sf]{subfig}
\usepackage{textcomp}
\usepackage{stfloats}
\usepackage{url}

\usepackage{verbatim}
\usepackage{graphicx}
\renewcommand{\vec}[1]{\bm{#1}} 
\def\BibTeX{{\rm B\kern-.05em{\sc i\kern-.025em b}\kern-.08em
    T\kern-.1667em\lower.7ex\hbox{E}\kern-.125emX}}
\usepackage{balance}
\begin{document}
\title{OSDM-MReg:  Multimodal Image Registration based One Step Diffusion Model}
\author{Xiaochen Wei,Weiwei Guo,~\IEEEmembership{Member,~IEEE},  Wenxian Yu~\IEEEmembership{Senior Member,~IEEE}, Feiming Wei,~\IEEEmembership{Member,~IEEE}, Dongying Li,~\IEEEmembership{Member,~IEEE}
\thanks{This work was supported by the National Natural Science Foundation of China  62071333. (\emph{Corresponding author: Weiwei Guo})
 Xiaochen Wei, Wenxian Yu, Dongying Li, Feiming Wei are with Shanghai Key Laboratory of Intelligent Sensing and Recognition, Shanghai Jiao Tong University, Shanghai 200240, China. 
 Weiwei Guo is with Center of Digital Innovation, Tongji University, Shanghai 200092, China.(email: weiweiguo@tongji.edu.cn) }}

\markboth{}%
{OSDM-MReg:  Multimodal Image Registration based One Step Diffusion Model}

\maketitle

\begin{abstract}
Multimodal remote sensing image registration aligns images from different sensors for data fusion and analysis. However, existing methods often struggle to extract modality-invariant features when faced with large nonlinear radiometric differences, such as those between SAR and optical images. To address these challenges, we propose OSDM-MReg, a novel multimodal image registration framework that bridges the modality gap through image-to-image translation. Specifically, we introduce a one-step unaligned target-guided conditional diffusion model (UTGOS-CDM) to translate source and target images into a unified representation domain. Unlike traditional conditional DDPM that require hundreds of iterative steps for inference, our model incorporates a novel inverse translation objective during training to enable direct prediction of the translated image in a single step at test time, significantly accelerating the registration process. After translation, we design a multimodal multiscale registration network (MM-Reg) that extracts and fuses both unimodal and translated multimodal images using the proposed multimodal fusion strategy, enhancing the robustness and precision of alignment across scales and modalities. Extensive experiments on the OSdataset demonstrate that OSDM-MReg achieves superior registration accuracy compared to state-of-the-art methods.
\end{abstract}

\section{Introduction}
\IEEEPARstart{M}{ultimodal} remote sensing image registration aligns images from different sensors—such as optical, SAR, infrared, and LiDAR—captured over the same area. Due to differences in sensing mechanisms, resolutions, and noise, these images vary significantly in geometry, texture, and radiometry, making registration highly challenging. Accurate alignment is crucial for downstream tasks including image fusion\cite{Ma2018InfraredAV,Zhang2023TransformerBC}, object detection\cite{Cao2023MultimodalOD,Belmouhcine2023MultimodalOD}, geo-localization\cite{Xiao2024STHNDH,Wang2020AttentionBasedRR}, and change detection\cite{Touati2020MultimodalCD,Luppino2020DeepIT}.

In recent years, multimodal image registration has become a prominent research topic, with numerous deep learning methods proposed\cite{zhao2021deep,zhang2023prise,zhangsparse}. Among them, iterative frameworks\cite{Cao2022IterativeDH,Cao2023RecurrentHE,Zhu2024MCNetRT,Xiao2024STHNDH} have shown promising performance. However, these methods typically focus on minimizing displacement loss at fixed control points while paying less attention to learning modality-invariant features. As a result, they often struggle when faced with large nonlinear radiometric differences, leading to reduced robustness and generalization across modalities.
To address these challenges, we propose OSDM-MReg, a novel multimodal registration framework based on image-to-image translation. Motivated by the success of diffusion models in image generation, we employ a conditional diffusion model (DDPM) to translate the source image into the target domain, thus narrowing the modality gap. However, traditional DDPMs are computationally expensive due to their multi-step inference. To overcome this, we introduce a Unaligned Target-Guided One-Step Conditional DDPM (UTGOS-CDM) that enables efficient one-step translation during inference. The main contributions of this work are as follows.
\begin{itemize}
    \item To eliminate the radiometric differences between cross-modal image pairs, we propose a novel multimodal image framework based on the image-to-image translation network, which utilizes the proposed unaligned target guided one step conditional diffusion model(UTGOS-CDM) to translate multimodal image pairs into one domain.
    \item To avoid a large number of iterations, UTGOS-CDM utilizes our proposed one-step strategy to train and inference and sets an unaligned target image as a condition to accelerate the generation of the low-frequency features in the translated image.
    \item To reduce the geometric errors and detail loss of the translated image that restricts the accuracy of multimodal image registration, we propose a novel dual-branch strategy to fuse the low-resolution features of the translated source images with the high-resolution features of the source images.
\end{itemize}

\section{Method}
\begin{figure*}
    \centering
    \includegraphics[width=\linewidth]{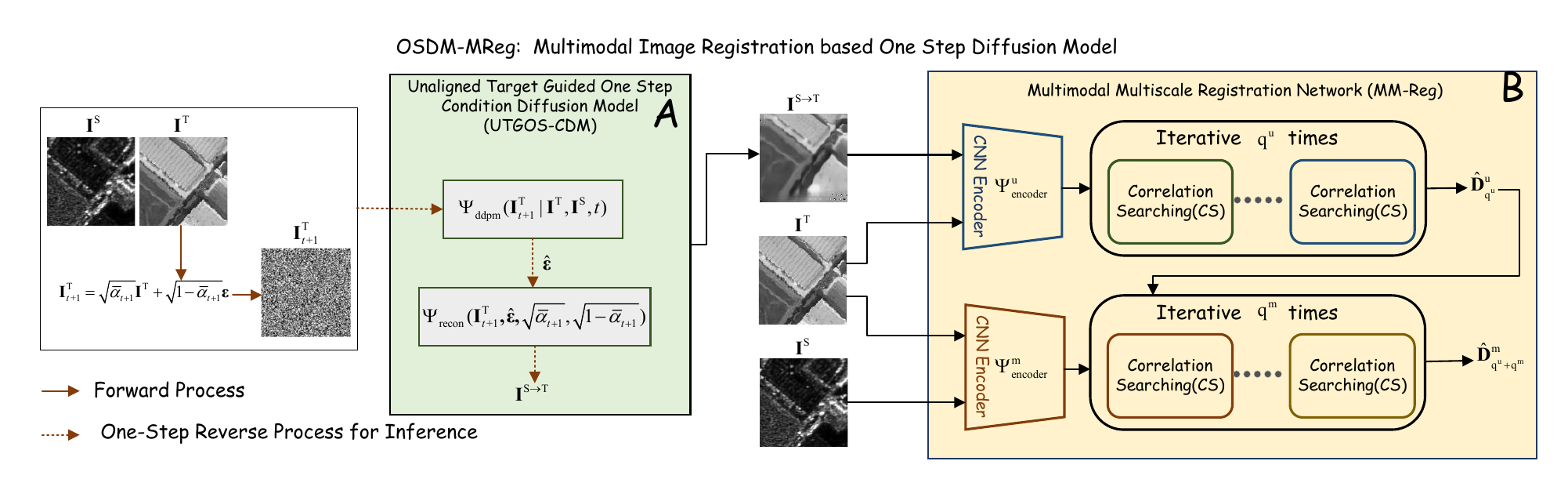}
    \caption{Overview of the proposed OSDM-MReg framework. 
The source image {\small $\vec{\mathrm{I}}^\mathrm{S}$} is first translated into the target domain via UTGOS-CDM, 
which employs a DDPM-based denoising network {\small $\Psi_{\mathrm{ddpm}}$} and a reconstruction module {\small $\Psi_{\mathrm{recon}}$} 
to generate {\small $\vec{\mathrm{I}}^{\mathrm{S} \rightarrow \mathrm{T}}$}. 
The unimodal pair {\small $\{\vec{\mathrm{I}}^{\mathrm{S} \rightarrow \mathrm{T}}, \vec{\mathrm{I}}^\mathrm{T}\}$} is then used 
to estimate the initial corner displacement {\small $\vec{\hat{\mathrm{D}}}_{\mathrm{q^u}}^\mathrm{u}$} via MM-Reg. 
Subsequently, the original multimodal pair {\small $\{\vec{\mathrm{I}}^\mathrm{S}, \vec{\mathrm{I}}^\mathrm{T}\}$} is utilized 
to predict the final displacement {\small $\vec{\hat{\mathrm{D}}}_{\mathrm{q^u}+\mathrm{q^m}}^\mathrm{m}$}, guided by the initial estimate.}
    \label{overview}
\end{figure*}
As shown in Fig. \ref{overview}, our multimodal image registration framework mainly consists of two parts. The first one is the \textbf{Unaligned Target Guided One Step Condition Diffusion Model(UTGOS-CDM)}, which is utilized to translate the source image {\small $\vec{\mathrm{I}}^\mathrm{S}$} from one domain into the other domain.  Source image {\small $\vec{\mathrm{I}}^\mathrm{S}$}, target image {\small $\vec{\mathrm{I}}^\mathrm{T}$}, and noise image {\small $\vec{\mathrm{I}}^\mathrm{T}_{t}$} are input into UTGOS-CDM to predict the noise {\small $\vec{\hat{{\mathrm{\varepsilon}}}}$}. And then the translated source image {\small $\vec{\mathrm{I}}^{\mathrm{S} \rightarrow \mathrm{T}}$} is generated by one-step reverse process{\small $\Psi_{\mathrm{recon}}(\vec{\mathrm{I}}_{t+1}^\mathrm{T},\vec{\hat{\mathrm{\varepsilon}}},\sqrt{\bar{\alpha}_{t+1}},\sqrt{1-\bar{\alpha}_{t+1}})$}
The other one is the \textbf{Multimodal Multiscale Image Registration Network(MM-Reg)}, which has two branches. The first branch is uimodal, which utilize the feature encoder {\small $\Psi_{\mathrm{encoder}}^\mathrm{u}$} to extract multiscale features of the unimodal image pairs {\small $\{\vec{\mathrm{I}}^{\mathrm{S}\rightarrow \mathrm{T}}, \vec{\mathrm{I}}^\mathrm{T}\}$},  and then input these feature into Correlation Searching(CS)\cite{Zhu2024MCNetRT} to obtain predicted displacements of four corners {\small $\vec{\hat{\mathrm{D}}}_{\mathrm{q^u}}^\mathrm{u}$} by iterating CS {\small $\mathrm{q}^\mathrm{u}$} times. The second branch is the multimodal branch. Be similar to the first branch, the cross-modality image pair {\small $\{\vec{\mathrm{I}}^\mathrm{S},\vec{\mathrm{I}}^\mathrm{T}\}$} is input into encoder {\small $\Psi_{\mathrm{encoder}}^\mathrm{m}$} to obtain multiscale features, and then CS utilizes these features and sets {\small $\vec{\hat{\mathrm{D}}}_{\mathrm{q^u}}^\mathrm{u}$} as initial estimation to predict the displacements of four corner {\small $\vec{\hat{\mathrm{D}}}_{\mathrm{q^u}+\mathrm{q^m}}^\mathrm{m}$} by iterating {\small $\mathrm{q^m}$} times.
\begin{figure}
    \centering
    \includegraphics[width=\linewidth]{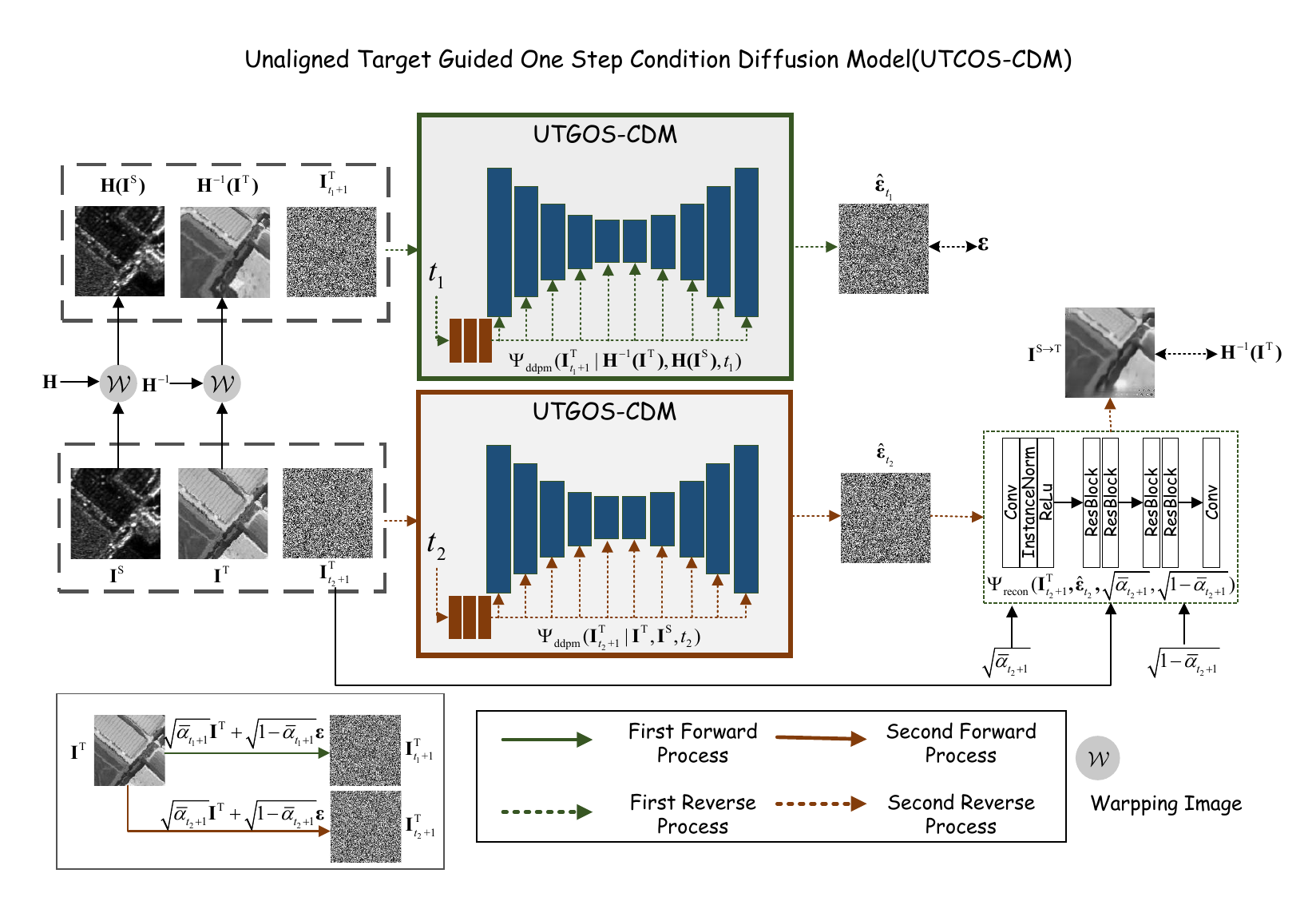}
    \caption{Overview of UTGOS-CDM. The model involves two forward and two reverse processes. Two noisy target images {\small $\vec{\mathrm{I}}^\mathrm{T}_{t_1+1}$} and {\small $\vec{\mathrm{I}}^\mathrm{T}_{t_2+1}$} are generated by adding Gaussian noise to {\small $\vec{\mathrm{I}}^\mathrm{T}$}. The first reverse process is conditioned on {\small ${\vec{\mathrm{H}}(\vec{\mathrm{I}}^\mathrm{S}), \vec{\mathrm{H}}^{-1}(\vec{\mathrm{I}}^\mathrm{T})}$}, and the second predicts the translated source image {\small $\vec{\mathrm{I}}^{\mathrm{S} \rightarrow \mathrm{T}}$} via one-step reconstruction. 
    }
    \label{figure_a}
\end{figure}
\subsection{Unaligned Target-Guided One Step Conditional Diffusion Model  (UTGOS-CDM)}
In recent years, conditional diffusion models have been widely adopted for multimodal image-to-image translation \cite{Li2022BBDMIT, Guo2024LearningSI, Xia2023DiffI2IED}. However, their direct application to multimodal image registration presents challenges. Specifically, these models typically require numerous iterative steps to translate an image from one modality to another, which significantly limits the efficiency of the registration process.
To address this limitation, we propose a novel Unaligned Target-Guided One-Step Conditional Diffusion Model (UTGOS-CDM), as illustrated in Fig.~\ref{figure_a}. During training, UTGOS-CDM incorporates newly designed forward and reverse processes that enable direct generation of the translated source image. As a result, during inference, UTGOS-CDM can synthesize the translated image {\small $\vec{\mathrm{I}}^{\mathrm{S} \rightarrow \mathrm{T}}$} in a single step.
In the following, we first describe the two forward processes, followed by a detailed explanation of the two corresponding reverse processes.
\subsubsection{Two Forwaed Processes}
As shown in Fig.~\ref{figure_a}, in two forward processes, UTGOS-CDM start with a target image {\small $\vec{\mathrm{I}}^\mathrm{T}$} and gradually add Gaussian noise {\small $\vec{\mathrm{\varepsilon}}$} to {\small $\vec{\mathrm{I}}^\mathrm{T}$} by {\small $t_1+1$} and {\small $t_{2}+1$} steps respectively, and generated two forward latent images {\small $\vec{\mathrm{I}}^\mathrm{T}_{t_1+1}$} and {\small $\vec{\mathrm{I}}^\mathrm{T}_{t_2+1}$} respectively, which are given by:
\begin{equation}
\label{fp}
\begin{aligned}
    &\vec{\mathrm{I}}^\mathrm{T}_{t_1+1}  = \sqrt{\bar{\alpha}_{t_1+1}} \vec{\mathrm{I}}^\mathrm{T} + \sqrt{1 - \bar{\alpha}_{t_1+1}}{\vec{\mathrm{\varepsilon}}}\\
    &\vec{\mathrm{I}}^\mathrm{T}_{t_2+1}  = \sqrt{\bar{\alpha}_{t_2+1}} \vec{\mathrm{I}}^\mathrm{T} + \sqrt{1 - \bar{\alpha}_{t_2+1}}{\vec{\mathrm{\varepsilon}}}\\
     &\bar{\alpha}_{t}= \prod_{s=1}^{s=t}1-\beta_{t}
\end{aligned}
\end{equation}
where {\small $\beta_t$} is a predefined positive constant.
The one forward process gradually perturbs {\small $\vec{\mathrm{I}}^\mathrm{T}$} to a latent variable with an isotropic Gaussian distribution. 
Another forward process gradually perturbs {\small $\vec{\mathrm{I}}^\mathrm{T}$} into a latent variable whose high-frequency features are contaminated by noise while the low-frequency features are preserved.
\subsubsection{Two Reverse Processes}
The two reverse processes are according to the two forward processes, as depicted in Fig.~\ref{figure_a}. The one reverse process is to predict the noise from the noise image {\small $\vec{\mathrm{I}}^\mathrm{T}_{t_1+1}$}, which is given by:
\begin{equation}
    \hat{\vec{\mathrm{\varepsilon}}}_{t_1} = \Psi_{\mathrm{ddpm}}(\vec{\mathrm{I}}_{t_1+1}^\mathrm{T},\vec{\mathrm{H}}^{-1}(\vec{\mathrm{I}}^\mathrm{T}),\vec{\mathrm{H}}(\vec{\mathrm{I}}^\mathrm{S}),t_1)
\end{equation}
where {\small $\vec{\mathrm{H}}$} is a homography transformation to align {\small $\vec{\mathrm{I}}^\mathrm{S}$} with {\small $\vec{\mathrm{I}}^\mathrm{T}$}, {\small $\vec{\mathrm{H}}^{-1}(\vec{\mathrm{I}}^\mathrm{T})$} and {\small $\vec{\mathrm{H}}(\vec{\mathrm{I}}^\mathrm{S})$} are condition, which can provide modality and geometry information, respectively.
Different with condition DDPM for image-to-image translation, our UTGOS-CDM utilizes the {\small $\vec{\mathrm{H}}^{-1}(\vec{\mathrm{I}}^\mathrm{T})$} to 
generate that there is not modality difference between the translated source image and the target image. 
For this reverse process, the estimated noise $\hat{\varepsilon}_{t_1}$ needs to be the same as the groundtruth $\varepsilon$ added in the forward process, as a result, the loss of this process is given by:
\begin{equation}
\mathcal{L}_{\mathrm{noise}} = \sum \vec{\mathrm{M}}^\mathrm{S}|\vec{\mathrm{\varepsilon}} - \vec{\hat{\mathrm{\varepsilon}}}_{t_1}|
\end{equation}
where {\small $\vec{\mathrm{M}}^\mathrm{S}\in\{0,1\}^{\mathrm{b} \times \mathrm{1}\times \mathrm{h} \times \mathrm{w}}$} is used to mask the padding pixels of {\small $\vec{\mathrm{H}}(\vec{\mathrm{I}}^\mathrm{S})$}.
In the training stage, the traditional condition diffusion models only need one reverse process, which set the aligned {\small $\vec{\mathrm{I}}^\mathrm{S}$} as a condition to predict noise from the latent variable {\small $\vec{\mathrm{I}}^\mathrm{T}_t$}.
In the inference stage, these models need large iterations to generate the translated source image, which greatly restricts the speed of image registration. 
To reduce time consumption, we propose a novel condition reverse process in training for one-step multimodal image-to-image  translation in inference, which is formulate as:
\begin{equation}
\begin{array}{l}
\vec{\mathrm{I}}^{\mathrm{S} \rightarrow \mathrm{T}} = \Psi_{\mathrm{recon}}(\vec{\mathrm{I}}_{t_2+1}^\mathrm{T},\vec{\hat{\varepsilon}}_{t_2},\sqrt{\bar{\alpha}_{t_2+1}},\sqrt{1-\bar{\alpha}_{t_2+1}})\\
\vec{\hat{\varepsilon}}_{t_2} = \Psi_{\mathrm{ddpm}}(\vec{\mathrm{I}}_{t_2+1}^\mathrm{T},\vec{\mathrm{I}}^\mathrm{T},\vec{\mathrm{I}}^\mathrm{S},t_2)
\end{array}
\end{equation}
Different with the first reverse process, in the second reverse process, we set {\small $\vec{\mathrm{I}}^\mathrm{T}$} and {\small $\vec{\mathrm{I}}^\mathrm{S}$} as modality and geometry condition, respectively. 
Guided by the low-frequency information of {\small $\vec{\mathrm{I}}^\mathrm{T}$} and high-frequency features of {\small $\vec{\mathrm{I}}^\mathrm{S}$}, 
the diffusion network learns to generate {\small $\vec{\mathrm{H}}^{-1}(\vec{\mathrm{I}}^\mathrm{T})$} from the noise image {\small $\vec{\mathrm{I}}_{t_2+1}$}.
Therefore, the translation loss of this reverse process to optimize the $\Psi_{\mathrm{ddpm}}$ and $\Psi_{\mathrm{recon}}$ is given by:
\begin{equation}
\mathcal{L}_{\mathrm{tran}} =  \sum \vec{\mathrm{M}}^\mathrm{T} |\vec{\mathrm{I}}^{\mathrm{S} \rightarrow \mathrm{T}} - \vec{\mathrm{H}}^{-1}(\vec{\mathrm{I}}^\mathrm{T})|
\end{equation}
where {\small $\vec{\mathrm{M}}^\mathrm{T}\{0,1\}^{\mathrm{b} \times 1\times \mathrm{h} \times \mathrm{w}}$} is used to mask the padding pixels of {\small $\vec{\mathrm{H}}^{-1}(\vec{\mathrm{I}}^\mathrm{T})$}.
\begin{figure}[!htbp]
    \centering
    \includegraphics[width=\linewidth]{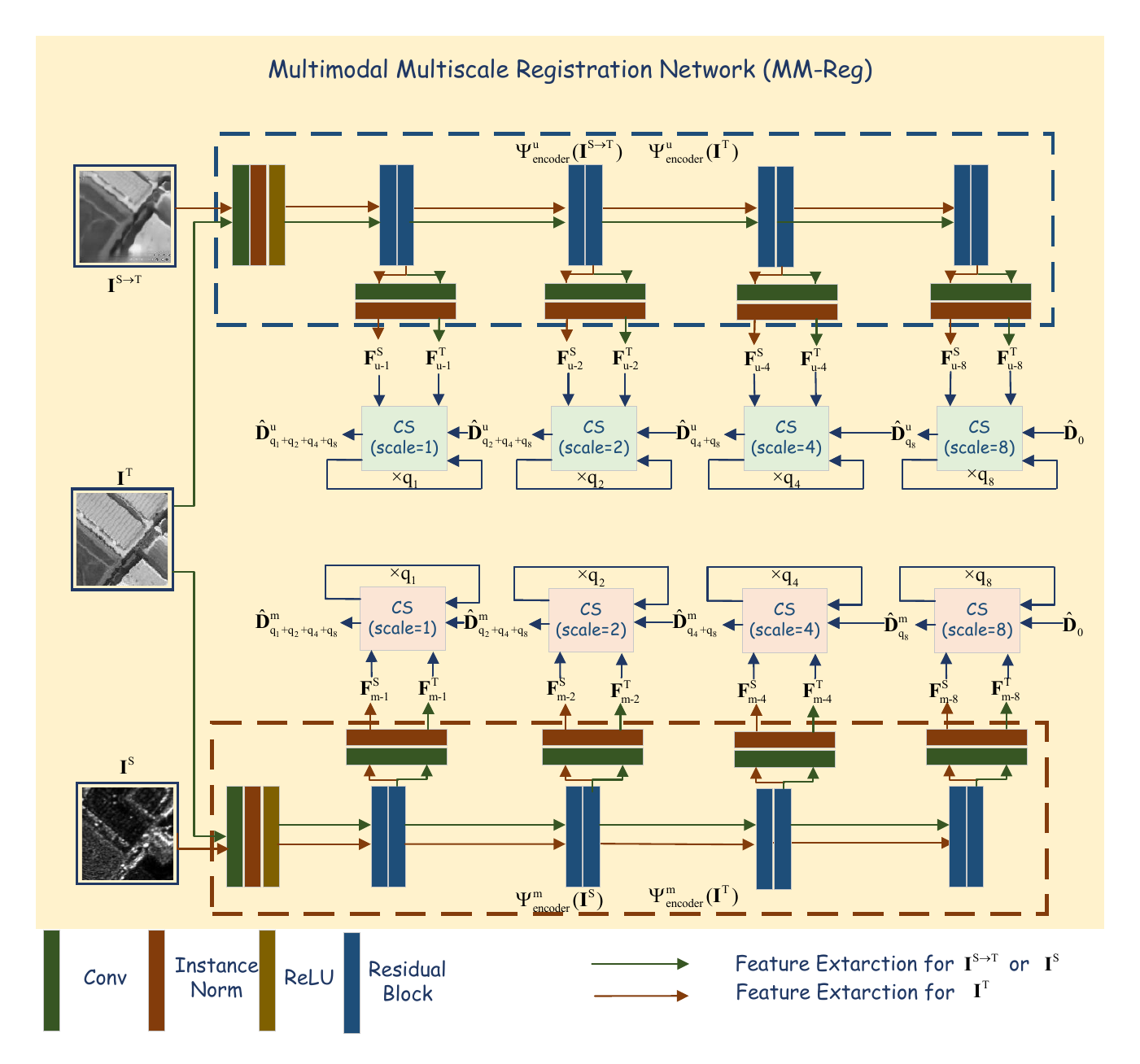}
    \caption{Training flow of MM-Reg. The framework contains two branches: (1) a unimodal branch with input {\small ${\vec{\mathrm{I}}^{\mathrm{S}\rightarrow \mathrm{T}}, \vec{\mathrm{I}}^\mathrm{T}}$}, and (2) a multimodal branch with input {\small $\{\vec{\mathrm{I}}^\mathrm{S}, \vec{\mathrm{I}}^\mathrm{T}\}$}. Both adopt multiscale iterative updates with 2 steps per scale, starting from {\small $\vec{\hat{\mathrm{D}}}_0 = \vec{\mathrm{0}}$}.}
    \label{figure_b}
\end{figure}
Therefore, the loss function $ \mathcal{L}_{\mathrm{diff}}$ for training UTGOS-CDM is calculated by:
\begin{equation}
    \mathcal{L}_{\mathrm{diff}} = \mathcal{L}_{\mathrm{noise}} + \mathcal{L}_{\mathrm{trans}}
\end{equation}

\begin{figure}[!htbp]
    \centering
    \includegraphics[width=\linewidth]{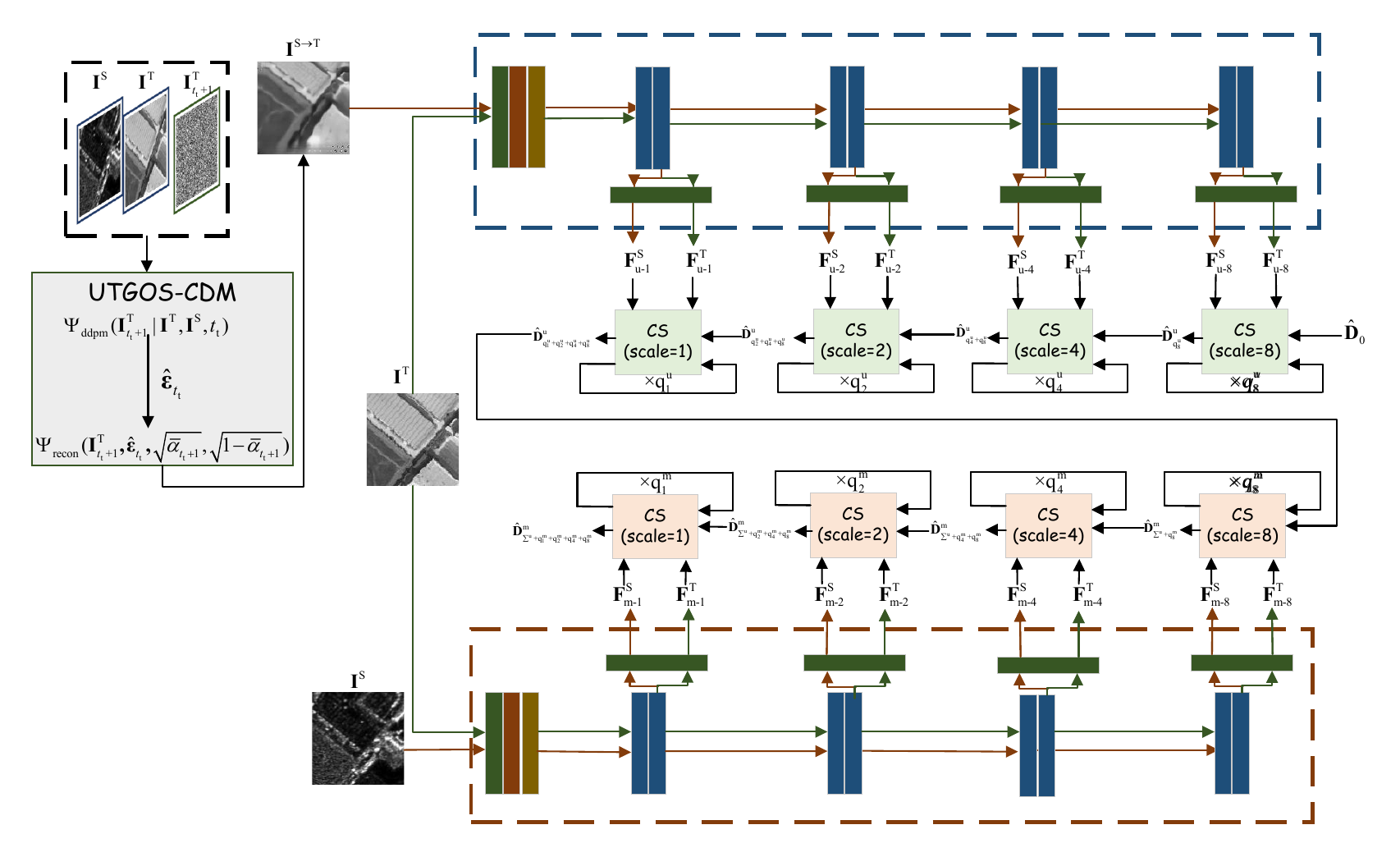}
    \caption{Test flowchart of the proposed OSDM-MReg. The unimodal prediction {\small $\vec{\hat{\mathrm{D}}}^\mathrm{u}_{\mathrm{q_1^u+q_2^u+q_4^u+q_8^u}}$} is used as the initial estimation for the multimodal branch to produce the final prediction {\small $\vec{\hat{\mathrm{D}}}^\mathrm{m}_{\mathrm{\sum + q_1^m+q_2^m+q_4^m+q_8^m}}$}. Stage weights are set to {\small $(2, 1, 0, 0)$} for the unimodal and {\small $(0, 1, 2, 2)$} for the multimodal branch during testing.}
    \label{test}
\end{figure}
\subsection{Multimodal Multiscale Registration Network (MM-Reg)}
To overcome the large appearance differences between multimodal images, we firstly utilizes pretrained UTGOS-CDM to translate {\small $\vec{\mathrm{I}}^\mathrm{S}$} into {\small $\vec{\mathrm{I}}^{\mathrm{S} \rightarrow \mathrm{T}}$}. Because there may be some blurred edges of objects in the translated source images {\small $\vec{\mathrm{I}}^{\mathrm{S} \rightarrow \mathrm{T}}$}, which will affect the network's ability to achieve high-precision registration performance. To address this issue, we propose a new strategy to fuse the registration results of {\small $\vec{\mathrm{I}}^{\mathrm{S} \rightarrow \mathrm{T}}$} and {\small $\vec{\mathrm{I}}^\mathrm{S}$}. Next, we will introduce the proposed MM-Reg in detail.
As shown in Fig. \ref{figure_b}, in training stage, MM-Reg is consist of two branches: the multimodal and unimodal branches, which utilize the multiscale feature maps {\small$\{\vec{\mathrm{F}}^\mathrm{S}_\mathrm{m-i}, \vec{\mathrm{F}}^\mathrm{T}_\mathrm{m-i} |\mathrm{i=1,2,4,8}\}$} and {\small $\{\vec{\mathrm{F}}^\mathrm{S}_\mathrm{u-i}, \vec{\mathrm{F}}^\mathrm{T}_\mathrm{u-i}|\mathrm{i=1,2,4,8}\}$} obtained by the multimodal encoder {\small $\Psi_{\mathrm{encoder}}^\mathrm{m}(\vec{\mathrm{I}}^\mathrm{S},\vec{\mathrm{I}}^\mathrm{T})$} and unimodal encoder {\small $\Psi_{\mathrm{encoder}}^\mathrm{u}(\vec{\mathrm{I}}^{\mathrm{S} \rightarrow \mathrm{T}},\vec{\mathrm{I}}^\mathrm{T})$}, respectively. {\small $\Psi_{\mathrm{encoder}}^\mathrm{m}$} and {\small $\Psi_{\mathrm{encoder}}^\mathrm{u}$} are feature extraction network in MCNet \cite{Zhu2024MCNetRT}.  Each branch starts with the lowest-resolution feature maps and ends with the feature maps that have the same resolution as the images. In each branch, we employ the multiscale correlation decoder module(CS) proposed by \cite{Zhu2024MCNetRT} to predict transformation parameters. Therefore, the loss for training registration network MM-Reg $\mathcal{L}_{\mathrm{reg}}$ is calculated by:
\begin{equation}
\begin{aligned}
    &\mathcal{L}_{\mathrm{reg}} = \mathcal{L}_{\mathrm{reg}}^\mathrm{u} + \mathcal{L}_{\mathrm{reg}}^\mathrm{m}\\
    &\mathcal{L}_{\mathrm{reg}}^\mathrm{bn} =  \sum_{j=0}^{j=\mathrm{N_{iter}}} (||\hat{\mathbf{D}}_{j}^\mathrm{bn}- \mathbf{D}||_1 + \mathcal{L}_{FGO}(||\Delta \hat{\mathbf{D}}_{j}^\mathrm{bn}- \mathbf{D}||_1))\\
    &\mathrm{bn}=\{\mathrm{u},\mathrm{m}\}
\end{aligned}
\end{equation}
where $\mathcal{L}_{FGO}$ is the Fine-grained Optimization Loss \cite{Zhu2024MCNetRT}, {\small $\mathrm{N_{iter}}$} is the number of iterations, {\small $\mathbf{P}$} denotes the groundtruth displacement of four corner points in source image {\small $\vec{\mathrm{I}}^\mathrm{S}$}.

\subsection{Inference}
As shown in Fig.~\ref{test}, in the testing stage, we firstly utilize the UTGOS-CDM to generate the translated source image {\small $\vec{\mathrm{I}}^{\mathrm{S}\rightarrow \mathrm{T}}$}, which is given by:
\begin{equation}
    \begin{aligned}
    &\vec{\mathrm{I}}^{\mathrm{S}\rightarrow \mathrm{T}} = \Psi_{\mathrm{recon}}(\vec{\mathrm{I}}_{t_\mathrm{t}+1}^\mathrm{T},\vec{\hat{\mathrm{\varepsilon}}}_{t_{\mathrm{t}}},\sqrt{\bar{\alpha}_{t_{\mathrm{t}+1}}},\sqrt{1-\bar{\alpha}_{t_{\mathrm{t}}+1}})\\
&\vec{\hat{\mathrm{\varepsilon}}}_{t_{\mathrm{t}}} = \Psi_{\mathrm{ddpm}}(\vec{\mathrm{I}}_{t_{\mathrm{t}+1}}^\mathrm{T},\vec{\mathrm{I}}^\mathrm{T},\vec{\mathrm{I}}^\mathrm{S},t_{\mathrm{t}})\\
   & \vec{\mathrm{I}}^\mathrm{T}_{t_{\mathrm{t}+1}} = \sqrt{\bar{\alpha}_{t_{\mathrm{t}+1}}} \vec{\mathrm{I}}^\mathrm{T} + \sqrt{1-\bar{\alpha}_{t_{\mathrm{t}+1}}} \vec{\mathrm{\varepsilon}}
    \end{aligned}
\end{equation}
where {\small $t_{\mathrm{t}}$} is the timestep selected in inference. 
Secondly, the image pair {\small $\{\vec{\mathrm{I}}^{\mathrm{S} \rightarrow \mathrm{T}},\vec{\mathrm{I}}^\mathrm{T}\}$} is input into the unimodal branch to obtain the prediction {\small $\vec{\hat{\mathrm{D}}}^\mathrm{u}_\mathrm{q_8^u+q_4^u+q_2^u+q_1^u}$}, which is set as initial prediction for multimodal branch with image pair {\small $\{\vec{\mathrm{I}}^\mathrm{S},\vec{\mathrm{I}}^\mathrm{T}\}$} to estimate the final prediction {\small $\vec{\hat{\mathrm{D}}}^\mathrm{m}_\mathrm{\sum +q_8^m+q_4^m+q_2^m+q_1^m}$}.
\section{Experiment and Results}
\subsection{Experimental Setup}
\subsubsection{Dataset}
OSdataset\cite{Xiang2020AutomaticRO} consists of 8044, 952, and 1696 pairs of {\small $256 \times 256$} aligned SAR and grayscale optical images for training, validation, and testing, respectively. The SAR and optical images are collected from GaoFen-3 and Google Maps, with a spatial resolution of 1m. Following the strategy proposed in DHN\cite{DeTone2016DeepIH}, we randomly generate unaligned cross-modal image pairs of size {\small $128 \times 128$} with a perturbation range of $[\pm 32]$.
We use three metrics: (1) ACE — mean Euclidean error of four corners; (2) AUC@k — proportion of samples with ACE below 3/5/7/10/15/20/25 pixels; (3) MACE — mean ACE over the dataset.
\subsubsection{Implementation Details}
We adopt a single NVIDIA A6000 to conduct all the experiments. We first train UTGOS-CDM with 3300K iterations.
For MM-Reg, we adopt the Adam optimizer and OneCycleLR scheduler with max learning rate $4e-4$ to train about 120K iterations.
\subsubsection{Compared Methods}
We compare our proposed method with other state-of-the-art deep learning methods for multimodal image Registration, which includes DHN\cite{DeTone2016DeepIH},MHN\cite{Le_CVPR_2020},IHN\cite{Cao2022IterativeDH}, MCNet\cite{Zhu2024MCNetRT}.
\subsection{Comparison}
\begin{figure}
    \centering
    \includegraphics[width=\linewidth]{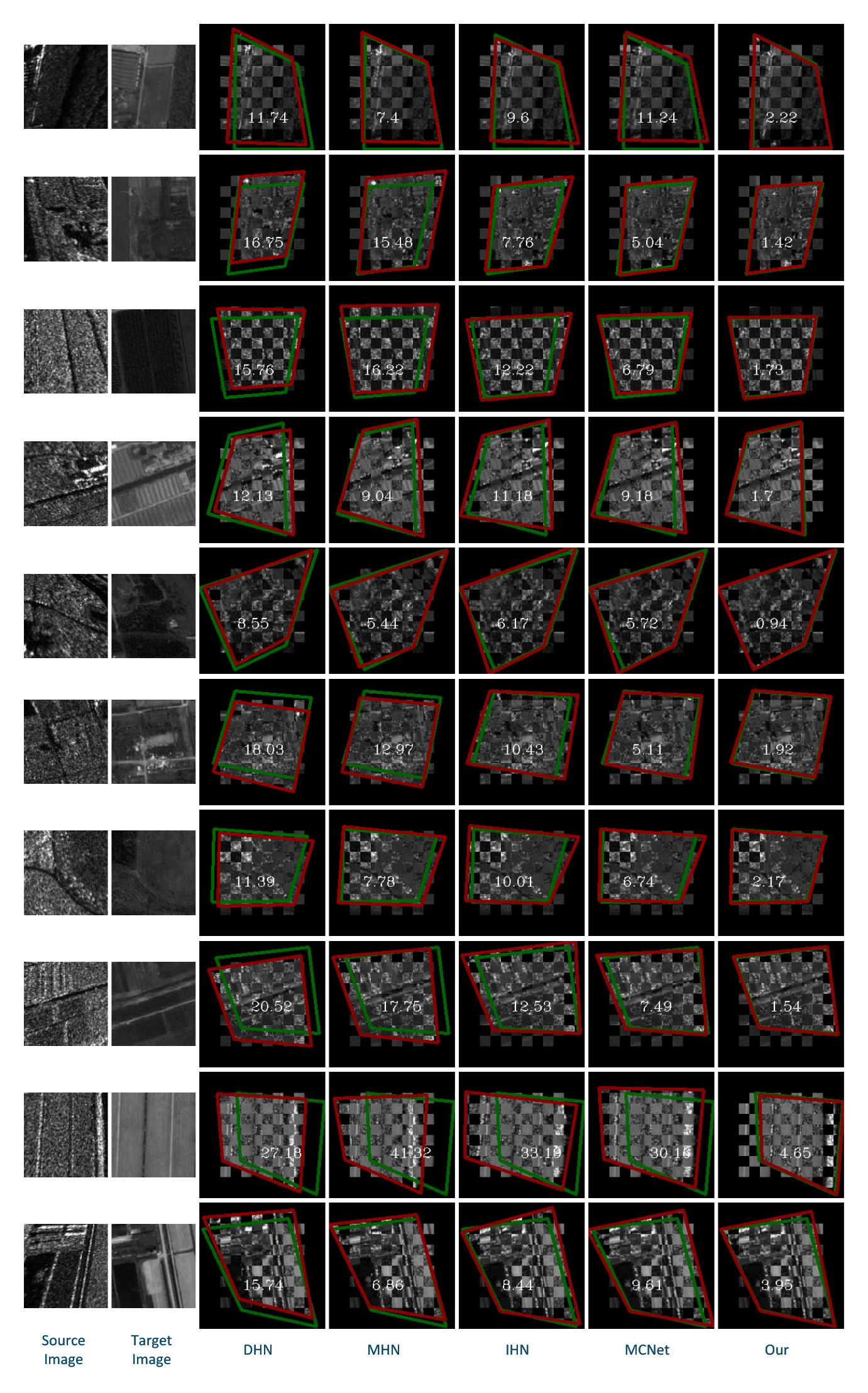}
    \caption{Qualitative homography estimation results. Green polygons denote the ground-truth homography deformation from source image to target image. Red polygons denote the estimated homography deformation using different methods on the target images.}
    \label{result}
\end{figure}

Compared with other multimodal registration tasks, SAR–optical registration is more challenging due to radiometric differences and speckle noise. Table.~\ref{table_gm} shows that our method achieves the best performance on OSdataset, with the lowest MACE (5.57) and a large margin in AUC metrics. Fig.~\ref{result} further demonstrates that our method maintains accurate alignment under severe texture and appearance differences. Benefiting from the UTGOS-CDM translation network, our approach effectively reduces modality gaps and speckle noise, enabling reliable registration even in low-texture regions.
\begin{table*}[!htbp]
\centering
\caption{Comparative results on OSdataset.Bold and underlined values indicate the best and second-best performance, respectively.}
\label{table_gm}
\begin{tabular}{l|rrrrrrr|r}
\hline
\textbf{Method} & AUC@3   & AUC@5 & AUC@7 & AUC@10 & AUC@15 &AUC@20 &AUC@25 &\textbf{MACE} \\ \hline
DHN                 & 0.2626 & 2.1117 &6.1917 &14.8662 &30.6268 &44.1862 &54.5005 & 11.4143 \\ \cline{1-9}
MHN                 & 0.4767 & 5.1595 &15.3100 & 31.7014 &50.9000 &62.2510  &69.5123 &7.6761 \\ \cline{1-9}
IHN                 &  0.6175 &5.6576 &15.1229 & 30.1761 &48.5415 & 59.9463 &67.4233  & 8.2570 \\ \cline{1-9}
MCNet               &\underline{0.8887} &\underline{7.4479} &\underline{18.6739} &\underline{35.1389} &\underline{53.2927} &\underline{63.9179} &\underline{70.7415} &\underline{7.4023} \\ \cline{1-9}
\textbf{OSDM-MReg}  & \textbf{4.6267} & \textbf{19.8763} & \textbf{34.7891} & \textbf{50.4504} & \textbf{64.9779} & \textbf{73.0075} & \textbf{78.0590}& \textbf{5.5716} \\
\hline
\end{tabular}
\end{table*}
\subsection{Ablation}
\begin{figure}
    \centering
    \includegraphics[width=80mm]{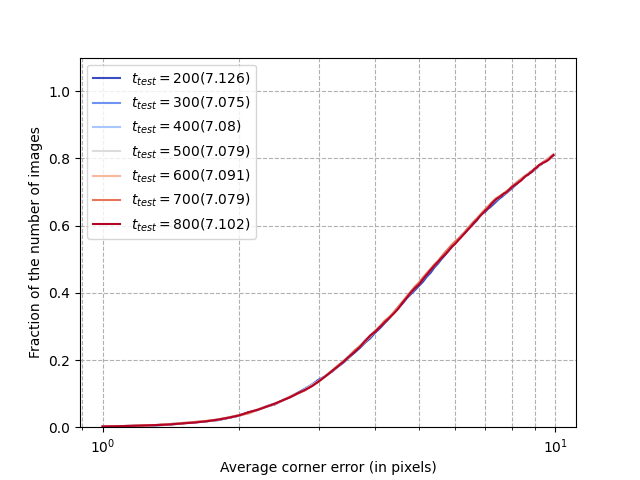}
    \caption{When time step {\small $t_\mathrm{t}=200,300,400,500,600,700,800$}, the average corner error of our OSDM-MReg on the validation dataset.}
    \label{diffusion}
\end{figure}
\subsubsection{Influence of Time Step {\small $t_\mathrm{t}$}}
We evaluate the influence of {\small $t_\mathrm{t}$} on OSDM-MReg using the OSdataset validation set with {\small $\mathrm{(q^u)=(2,2,2,2)}$} and {\small $\mathrm{(q^m)=(0,0,0,0)}$}. As shown in Fig., performance is insensitive to $t_\mathrm{t}$, so we set $t_\mathrm{t}=500$ for testing, the midpoint of [200,800].

 \subsubsection{Ablation of unimodal and multimodal branch}
\begin{table}[!htbp]
 \caption{Comparative results on validation set of OSdataset, when we set different {\small $\mathrm{(q_8^u,q^u_4,q^u_2,q^u_1),(q_8^m,q^m_4,q^m_2,q^m_1)}$} for OSDM-MReg.}
 \label{a_um}
\centering
\begin{tabular}{c|c}
\hline
{\small $\mathrm{(q_8^u,q_4^u,q_2^u,q_1^u)}$}
{\small $\mathrm{(q_8^m,q_4^m,q_2^m,q_1^m)}$}    & MACE $\downarrow$\\ \hline
(0,0,0,0),(2,2,2,2)  & 7.793              \\ \hline
(1,0,0,0),(1,2,2,2)  & 7.257              \\ \hline
(2,0,0,0),(0,2,2,2)  & 6.784               \\ \hline
(2,1,0,0),(0,1,2,2)  & \textbf{6.480}       \\ \hline
(2,2,0,0),(0,0,2,2)  & 6.835               \\ \hline
(2,2,1,0),(0,0,1,2)  & 6.879                 \\ \hline
(2,2,2,0),(0,0,0,2)  & 7.091                 \\ \hline
(2,2,2,1),(0,0,0,1)  & 7.075                  \\ \hline
(2,2,2,2),(0,0,0,0)  & 7.079                    \\ \hline
\end{tabular}
\end{table}
In testing, we fuse unimodal and multimodal branches with parameters {\small $\mathrm{\{q^u\}}$} and {\small $\mathrm{\{q^m\}}$}. As shown in Table~\ref{a_um}, unimodal cues improve over the multimodal-only baseline, but geometric errors cause MACE to first drop then rise; thus we set {\small $\mathrm{(2,1,0,0)}$} and {\small $\mathrm{(0,1,2,2)}$}.

\section{Conclusion}
In this paper, we presented a novel multimodal image registration framework, OSDM-MReg, which leverages image-to-image translation to effectively address the radiometric differences between cross-modal image pairs. By introducing the Unaligned Target-Guided One-Step Conditional  Diffusion Model (UTGOS-CDM), we successfully mapped multimodal images into a unified domain, eliminating modality disparities. The proposed one-step generation strategy accelerated the image translation process, avoiding the need for extensive iterations required by traditional methods. The dual-branches fusion strategy combined low-resolution features from the translated source image with high-resolution features from the original source image, effectively minimizing geometric errors and enhancing the registration accuracy. Experiments demonstrated that OSDM-MReg outperforms existing methods in terms of accuracy, particularly in SAR-optical image registration tasks. 
\bibliographystyle{ieeetr}
\bibliography{bibtex}

\end{document}